\documentclass[11pt, a4paper, logo, twocolumn, internal, copyright, nonumbering]{deepmind}

\pdftrailerid{redacted}

\makeatletter
\renewcommand\bibentry[1]{\nocite{#1}{\frenchspacing\@nameuse{BR@r@#1\@extra@b@citeb}}}
\makeatother

\usepackage{kantlipsum, lipsum}
\usepackage{dsfont}
\usepackage{dm-colors}

\usepackage{mathtools}
\usepackage{dsfont}
\usepackage[colorinlistoftodos]{todonotes}
\usepackage{booktabs}
\usepackage{xfrac}
\usepackage{bbm}

\usepackage{algorithm}
\usepackage{algorithmicx}
\usepackage{algpseudocode}

\usepackage[most]{tcolorbox}
\usepackage{xparse}
\usepackage{lipsum}
\usepackage{changepage}
\usepackage{enumitem}

\newcommand{\squishlist}{
   \begin{list}{$\bullet$}
    { \setlength{\itemsep}{0pt}      \setlength{\parsep}{3pt}
      \setlength{\topsep}{3pt}       \setlength{\partopsep}{0pt}
      \setlength{\leftmargin}{1.5em} \setlength{\labelwidth}{1em}
      \setlength{\labelsep}{0.5em} } }

\newcommand{\squishlisttwo}{
   \begin{list}{$\bullet$}
    { \setlength{\itemsep}{0pt}    \setlength{\parsep}{0pt}
      \setlength{\topsep}{0pt}     \setlength{\partopsep}{0pt}
      \setlength{\leftmargin}{2em} \setlength{\labelwidth}{1.5em}
      \setlength{\labelsep}{0.5em} } }

\newcommand{\squishend}{
    \end{list}  }

\newcommand{\two}{(2)}

\DeclareMathAlphabet{\mathpzc}{OT1}{pzc}{m}{n}

\usepackage[authoryear, sort&compress, round]{natbib} 

\graphicspath{{figures/}}

\title{Motion Tracking with Muscles:\\Predictive Control of a Parametric Musculoskeletal Canine Model}

\correspondingauthor{vlabarbera@rvc.ac.uk}


\author[1, 2]{Vittorio La Barbera}
\author[1]{Steven Bohez}
\author[1]{Leonard Hasenclever}
\author[1]{Yuval Tassa}
\author[2]{John R. Hutchinson}

\affil[1]{DeepMind}
\affil[2]{Royal Veterinary College}

\begin{abstract}
We introduce a novel musculoskeletal model of a dog, procedurally generated from accurate 3D muscle meshes. Accompanying this model is a motion capture-based locomotion task compatible with a variety of control algorithms, as well as an improved muscle dynamics model designed to enhance convergence in differentiable control frameworks. We validate our approach by comparing simulated muscle activation patterns with experimentally obtained electromyography (EMG) data from previous canine locomotion studies. This work aims to bridge gaps between biomechanics, robotics, and computational neuroscience, offering a robust platform for researchers investigating muscle actuation and neuromuscular control. We plan to release the full model along with the retargeted motion capture clips to facilitate further research and development.
\end{abstract}

\begin{document}

\maketitle


\section{Introduction}

\begin{figure*}[t]
\centering
\includegraphics[width=4cm]{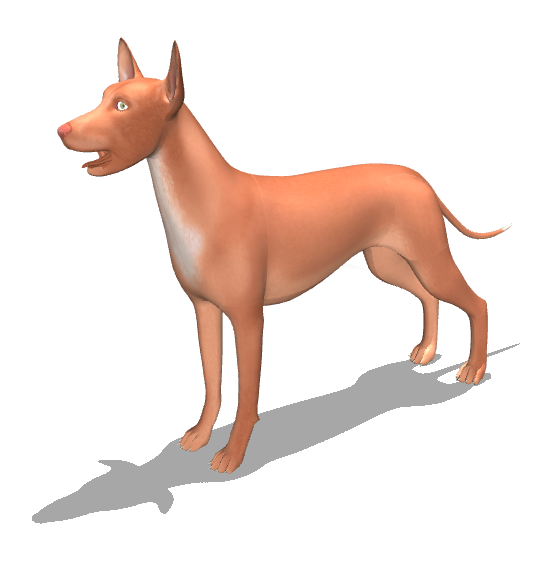}
\hfill
\includegraphics[width=4cm]{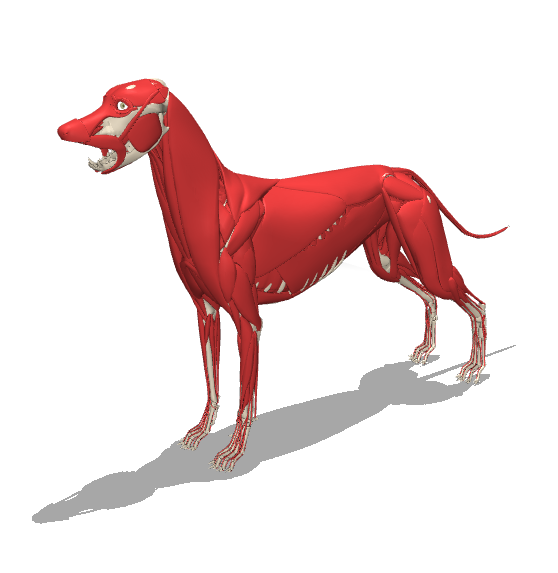}
\hfill
\includegraphics[width=4cm]{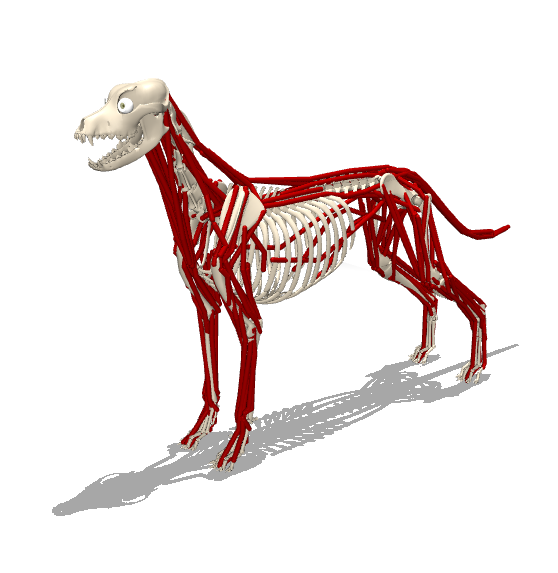}
\end{figure*}

\begin{figure*}[t]
\centering
\includegraphics[width=5.2cm]{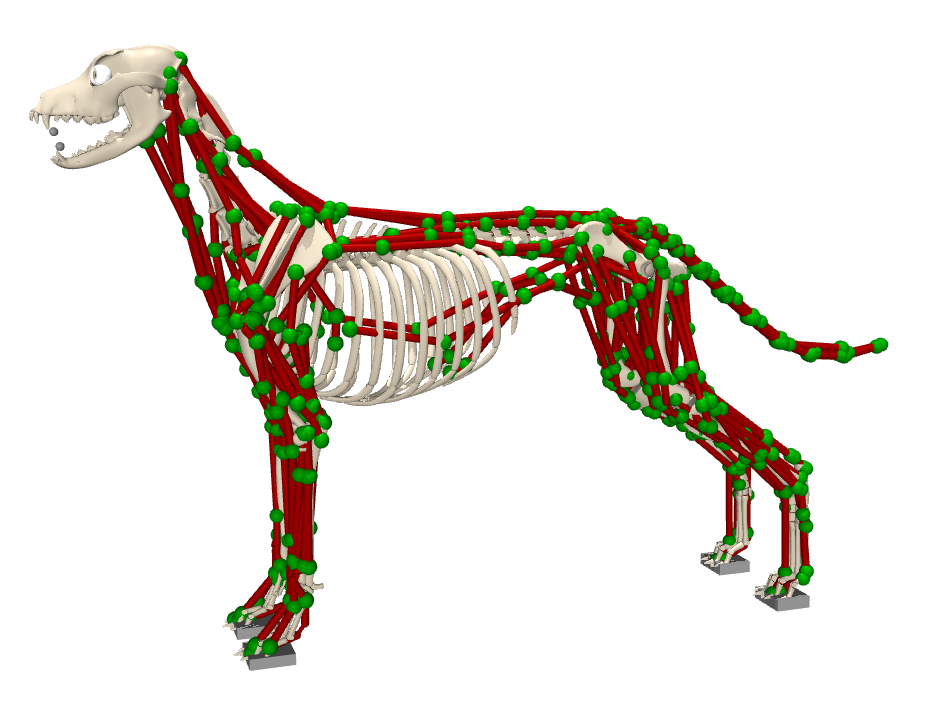}
\hspace{0.12cm}
\hfill
\includegraphics[width=5.2cm]{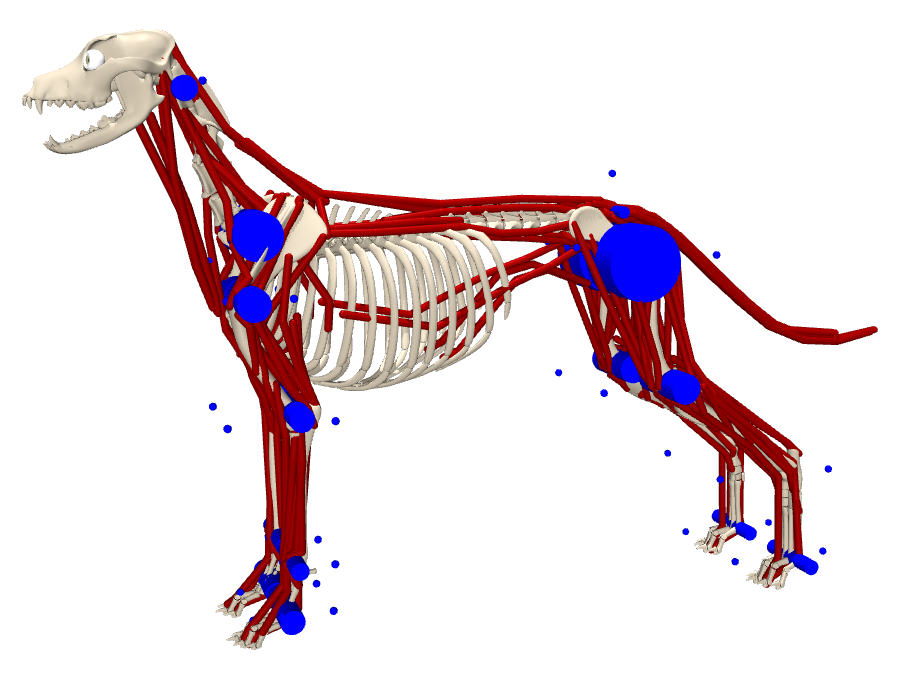}
\hfill
\includegraphics[width=4.5cm]{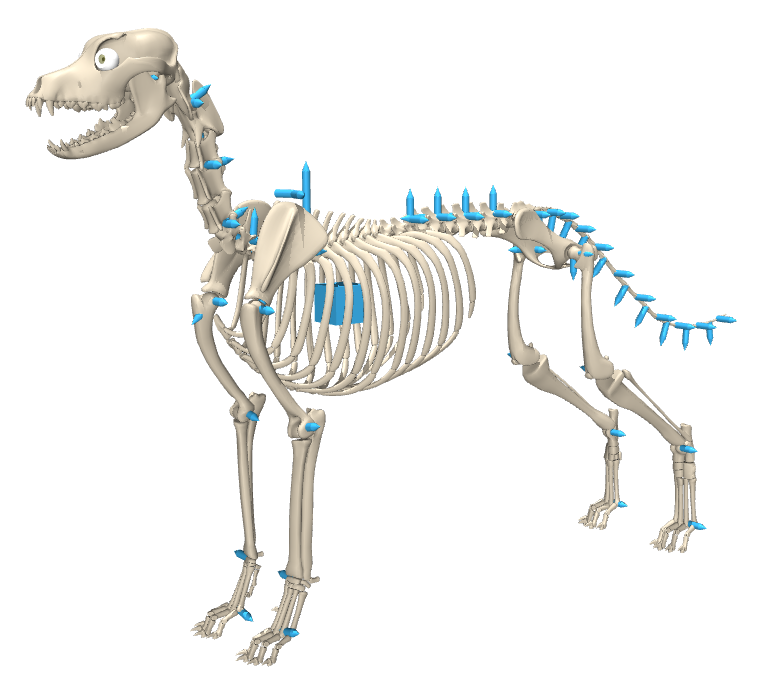}
\caption[The proposed model]{This image shows the proposed musculoskeletal model. In the top row we show the skins available in the model and the muscles used to control the dog. In the bottom row we highlight what makes this model complex: all the muscle way points in green, the wrapping geometries in blue and the degrees of freedom in light blue.}
\label{fig:model}
\end{figure*}

Musculoskeletal systems comprise highly intricate networks of muscles, tendons, ligaments, bones, and joints, playing a fundamental role in enabling animals to execute diverse and complex movements. Understanding the mechanical behavior and dynamics of these interconnected systems is particularly challenging due to their complexity and variability. Accurate physics-based simulations have emerged as valuable tools to overcome these challenges, allowing detailed studies of musculoskeletal dynamics in an animal welfare-neutral manner while providing comprehensive insights into muscle interactions.

Musculoskeletal modeling is an inherently multidisciplinary field, integrating principles from anatomy, physiology, biomechanics, and computational modeling. It has been extensively investigated over the past decades in biomechanics research \citep{zajac1989muscle}, proving invaluable for exploring phenomena such as movement coordination, posture control, and the impacts of injuries and diseases. These insights have widespread applications across multiple disciplines: biologically, they elucidate how animals achieve their diverse range of gaits; in graphics and animation, they facilitate the creation of realistic digital twins used in movies and video games; and in sports science, they enhance injury prevention and recovery strategies.

To control musculoskeletal systems or torque-actuated characters, researchers predominantly employ two methodologies: traditional model-based trajectory optimization and modern reinforcement learning (RL). Central to many contemporary studies, including this work, is the application of these techniques to control sophisticated, physics-based models, aiming to replicate natural and complex movements. Reinforcement learning involves agents interacting with an environment to maximize cumulative rewards, offering adaptability and efficiency in generating natural locomotion \citep{Sutton1998Reinforcement}.

In this study, we specifically focus on quadrupedal locomotion, the movement strategy employed by animals standing and moving on four limbs. Quadrupedal locomotion encompasses a diverse set of gaits, including lateral and diagonal sequence walking, pacing, trotting, galloping, cantering, as well as complex actions such as jumping, turning, sitting, and idling \citep{hilderbrand2015Adaptive, Hildebrand1968Symmetrical}. These movements require sophisticated coordination among limbs to achieve efficient and stable locomotion.

The domestic dog was selected as the primary test subject for this research due to several compelling reasons. Dogs exhibit considerable morphological diversity across breeds, presenting substantial variations in size, shape, and body mass. While the presented work focuses on a single model, this inherent variability underscores the potential to generalize findings across a wide range of physical forms in future studies. Additionally, dogs frequently experience musculoskeletal disorders analogous to those found in humans, such as hip dysplasia and arthritis, making them valuable models for medical and veterinary studies. The extensive existing literature on canine anatomy, physiology, and biomechanics further facilitates comparative studies and validation of simulation results \citep{Walter2009Rapid, Usherwood2007Mechanics, carrier1999Dynamic}.

To effectively investigate canine biomechanics, we introduce a novel, highly detailed musculoskeletal model derived from precise 3D muscle geometries and implemented using the efficient physics engine MuJoCo \citep{todorov2012mujoco}. Figure \ref{fig:model} illustrates the developed model, showcasing bone geometry, skin surfaces, and tendon routing.

This model supports high-fidelity simulations of canine locomotion, offering a robust computational platform to examine neuromuscular control strategies and biomechanical function. To generate these simulations and investigate control, we employ optimization methods where the primary objective is to track reference motion data derived from canine locomotion. This tracking-focused optimization yields estimates of critical muscle functions, such as excitation patterns, which we then compare against existing experimental data.

By bridging experimental observations and computational simulations, this research advances our understanding of canine biomechanics, with practical implications for veterinary medicine, rehabilitation therapies, and bio-inspired robotics. The proposed model can further support studies of pathological gaits, optimization of prosthetic devices, and the development of improved control strategies for quadrupedal robots.

Finally, investigating canine locomotion provides valuable insights into the design and control of quadrupedal robotic systems. Dogs naturally traverse varied terrains, efficiently adapting their gait patterns to uneven, slippery, and rugged surfaces. Understanding these biological principles enables the development of adaptive, agile, and energy-efficient robotic control systems, contributing significantly to advancements in robotics.

This paper primarily contributes a novel, high-fidelity musculoskeletal dog model derived from 3D muscle meshes, along with an improved muscle dynamics model and a motion-capture-based control task. We demonstrate its utility by generating simulations that track reference motions and validate muscle activation patterns against experimental EMG data. We further intend to release the model and associated data, providing a valuable resource for the biomechanics, robotics, and computational neuroscience communities.

\section{Related Works}
\label{related-work}

Research on canine biomechanics encompasses various fields, notably focusing on locomotor muscle function. Electromyography (EMG), when combined with motion capture and biomechanical modeling, has significantly advanced the understanding of muscle dynamics in canine locomotion.

\citet{Goslow1981ElectricalAA} pioneered EMG studies by analyzing electrical activity and muscle length changes in canine limbs during walking, trotting, and galloping. They demonstrated that muscles engage in diverse contraction modes (concentric, eccentric, and isometric), revealing specialized functions beyond simple force production. \citet{Deban2012Activity} expanded on these insights by examining extrinsic limb muscles across different gaits, consistently finding conserved roles such as limb protraction, retraction, and trunk stabilization. \citet{cullen2017magnitude} further investigated muscle activation patterns specifically in agility dogs, noting significantly elevated muscle activity during agility tasks, which correlated with increased injury risk.

Several simulation studies have developed musculoskeletal models to understand canine biomechanics better. \citet{stark2021three} presented a comprehensive Beagle musculoskeletal model, using OpenSim \citep{Ajay2018OpenSim} to simulate muscle activations and validate these against literature-derived EMG data. Their model showed generally good agreement, though discrepancies occurred for some muscles due to anatomical variability or modeling assumptions.

\citet{Brown2020development} focused on pelvic and hindlimb muscle activation in Dachshunds, clearly differentiating stance and swing phase muscle roles. Although direct EMG comparisons were not performed, their results were validated with literature-derived data. \citet{Ellis2018Limb} studied muscle dynamics during sit-to-stand transitions in Greyhounds, identifying early-phase peak muscle activations reflecting high initial mechanical demands. Additionally, \citet{ALIZADEH2017AnEMG} investigated cervical biomechanics, combining EMG and motion capture. They identified discrepancies in internal muscle force predictions, suggesting possible overestimations by simulation models.

Controlling musculoskeletal models in simulations introduces further complexity due to inherent nonlinear and delayed muscle dynamics. Detailed analyses by \cite{Peng2016Learning} underscored the significance of actuation choice in reinforcement learning (RL), emphasizing the necessity for detailed and robust models to advance novel control algorithms. To facilitate algorithmic advancements, various communities have contributed high-quality models. Examples include an ostrich musculoskeletal model developed by \citet{laBarbera21OstrichRL}, and a human arm model released by \cite{caggiano22MyoSuite}, both supporting realistic and detailed control studies.

In the graphics community, \citet{He2018ModeAdaptive} proposed a kinematic dog model generating diverse locomotion patterns but lacking dynamic muscle activation and ground reaction forces estimation. \citet{Lee2019ScalableMH} was the first to propose a reinforcement learning agent able to control a complex musculoskeletal model with 346 muscles. To control such a complex model reference trajectories were used to produce natural behaviours. Moreover, the study applied an inductive bias using two neural networks for the agent, breaking down the control into two levels: trajectory mimicking and muscle coordination. \citet{Geijtenbeek2013} optimized muscle routing combined with PD controllers through evolutionary strategies. 

Recently, reinforcement learning frameworks like Differential Extrinsic Plasticity (DEP) by \citet{schumacher2023deprl}, and structured noise injection methods by \citet{chiappa2023latentexplorationreinforcementlearning, plappert2018parameterspacenoiseexploration}, have significantly improved performance on complex musculoskeletal tasks. Another promising strategy to manage the high dimensionality of musculoskeletal control is to exploit muscle synergies—coordinated activations of muscle groups that serve as lower-dimensional modules for movement. By using synergy-inspired action spaces, recent approaches have achieved more efficient and generalizable control in complex musculoskeletal models. For example, \citet{berg2023sargeneralizationphysiologicalagility} present the Synergistic Action Representation (SAR), which autonomously discovers low-dimensional muscle synergy patterns from simpler tasks and uses them as an action basis for learning more complex behaviors.

Overall, prior studies reveal important insights and remaining limitations in canine musculoskeletal modeling and control methods. Our research aims to integrate detailed anatomical accuracy and advanced control frameworks, bridging the gap between experimental biomechanics, simulation-based validation, and control algorithm development.

\section{Methods}

\subsection{Model Creation}

\begin{figure*}[t]
\centering
\includegraphics[width=16cm]{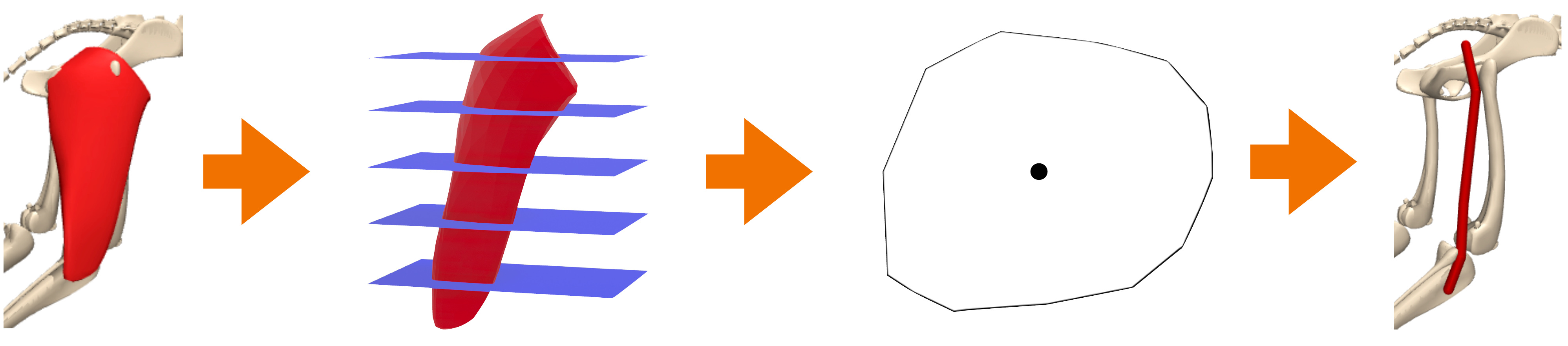}
\caption[The proposed model]{The steps necessary to estimate the line of action of a muscle. Starting from a muscle mesh shown in the left then slicing it along its longitudinal axis and obtaining a centroid for each slice. Lastly we pick the least amount of centroids for the line of action displayed in the right.}
\label{fig:pipeline}
\end{figure*}

In this section we describe how we built the proposed musculoskeletal model using MuJoCo \citep{todorov2012mujoco}.
Before diving in the details it is necessary to explain how muscles are modelled geometrically.
In MuJoCo muscles are not volumetric, hence are approximated by a single line of action and contract following the muscle dynamics. Muscles are attached to at least two bones, using two sites at the extremities, called the origin and the insertion sites. Other sites, called waypoints or via points, and wrapping geometries, can also be specified to define elaborate muscle routings. Waypoints are sites through which muscles must pass, which are useful to maintain anatomically realistic 3D paths. Wrapping geometries as the name suggests are geometric primitives such as spheres or cylinders that muscles must wrap around, these provide a useful way to prevent muscles from penetrating bones or other structures.
More details can be found in the official \href{https://mujoco.readthedocs.io/en/stable/modeling.html#cmuscle}{MuJoCo documentation}.

A detailed torque actuated model of a Pharaoh Dog, with all the skeletal geometries, was already available in the \texttt{dm\_control} suite made by \href{https://www.turbosquid.com/Search/Artists/leo3Dmodels}{leo3Dmodels} . We extended the model by adding 133 muscles to it, using a set of 3D muscle meshes also already available in the \texttt{dm\_control} suite.

Determining a muscle line of action (LoA) from a 3D mesh is challenging because muscles have complex, non-linear paths that wrap around bones and soft tissues, making a simple straight-line approximation insufficient. Accurate extraction requires careful consideration of anatomical constraints, dynamic movement, and computational efficiency, often relying on centroidal paths, via-points, or wrapping algorithms. A well-defined LoA is critical for estimating muscle forces and joint torques, impacting applications in biomechanics, rehabilitation, and bio-inspired robotics. We propose a new simple algorithm that follows the longitudinal axis of the muscle mesh to create a LoA. 

Obtaining a muscle line of action from a 3D mesh has been studied before in the biomechanics community by \cite{demuth2022Three-dimensional}. They estimate the line of action based on the muscle's 3D volume, which incorporates the entire muscle mass and the tendon structure. Their method looks at the internal distribution of muscle fibres within the volume and uses that to determine how forces are applied. Demuth's approach can be computationally expensive since it requires volumetric data and detailed reconstruction of muscle geometry. In contrast, our proposed approach estimates the line of action by focusing on 2D slices and centroids of the muscle's path, not the volume, which is more computationally efficient, especially when dealing with complex meshes. 

We follow a procedural algorithm that processes each muscle independently.
We first slice the 3D mesh along the longitudinal axis of the muscle with equally separated planes, with the number of slices proportional to the length of the muscle along that axis. Each slice will contain a contour of the muscle, from which we then compute the centroid and store it as a 3D point. We used Trimesh \citep{trimesh} to compute the slicing and centroids. We then iterate through the centroids and search for the bone in the skeleton. We accelerate the search of the closest geometry by using kd-trees \citep{Bentley1975kdtree} for each bone. We then add the centroid to the tendon path if it satisfies one of the following conditions:
\begin{itemize}
    \item The centroid is either the first or the last (i.e. if it's either an origin or insertion point).
    \item The distance between the current centroid and the one previously added is above a certain threshold (we want a path that geometrically resembles the muscle mesh, so this is done to ensure uniform sampling along the longitudinal axis), depending on the muscle (muscles that span multiple joints and therefore pass by more bones are generally longer and might have a more complex path, hence requiring a lower threshold) this threshold varies between 5 and 15 cm.
    \item The closest bone to the current centroid is different from the previous bone and the distance between the current centroid and the one previously added is above 5 cm. 
\end{itemize}
The algorithm is outlined in a more compact way in the Supplementary Materials Sec.~\ref{line of action} and shown in Fig.~\ref{fig:pipeline}. 

After the LoA is created, we estimate the peak active ($F_0$) force for each muscle.
The output of the Force-Length-Velocity (FLV) function is the scaled muscle force. The scaled force is then multiplied by a muscle-specific constant $F_0$ to obtain the actual force:

\begin{equation}
    F_{\text{actuator}} = - \text{FLV}(L, V, \text{act}) \cdot F_0
\end{equation}

MuJoCo roughly estimates $F_0$ by dividing a scale constant by the acceleration from unit force in the initial position.

\begin{equation}
    F_0 = \frac{\text{scale}}{\text{actuator\_acc0}}
\end{equation}

The quantity $actuator\_acc0$ is pre-computed by the model compiler. It is the norm of the joint acceleration caused by unit force acting on the actuator transmission. Intuitively, scale determines how strong the muscle is “on average” while its actual strength depends on the geometric and inertial properties of the entire model.

An alternative approach involves estimating the peak muscle force ($F_0$) based on the available muscle geometry. This method relies on calculating the physiological cross-sectional area (PCSA), which is defined as the muscle volume divided by the fibre length. However, deriving accurate data from geometry alone presents significant challenges. One key issue is that the muscle meshes do not distinctly separate the biological tendon from the muscle tissue. As a result, the only feasible option is to include the entire volume of both muscle and tendon in the calculations. We tested these estimated forces in simulation, but the model failed to accurately track the reference motion, most likely because the computed force values deviated too much from expected biological values, leading to unrealistic muscle activations and impaired motion execution.

\subsection{Motion Capture}

\begin{figure*}
    \centering
    \includegraphics[width=0.3\linewidth]{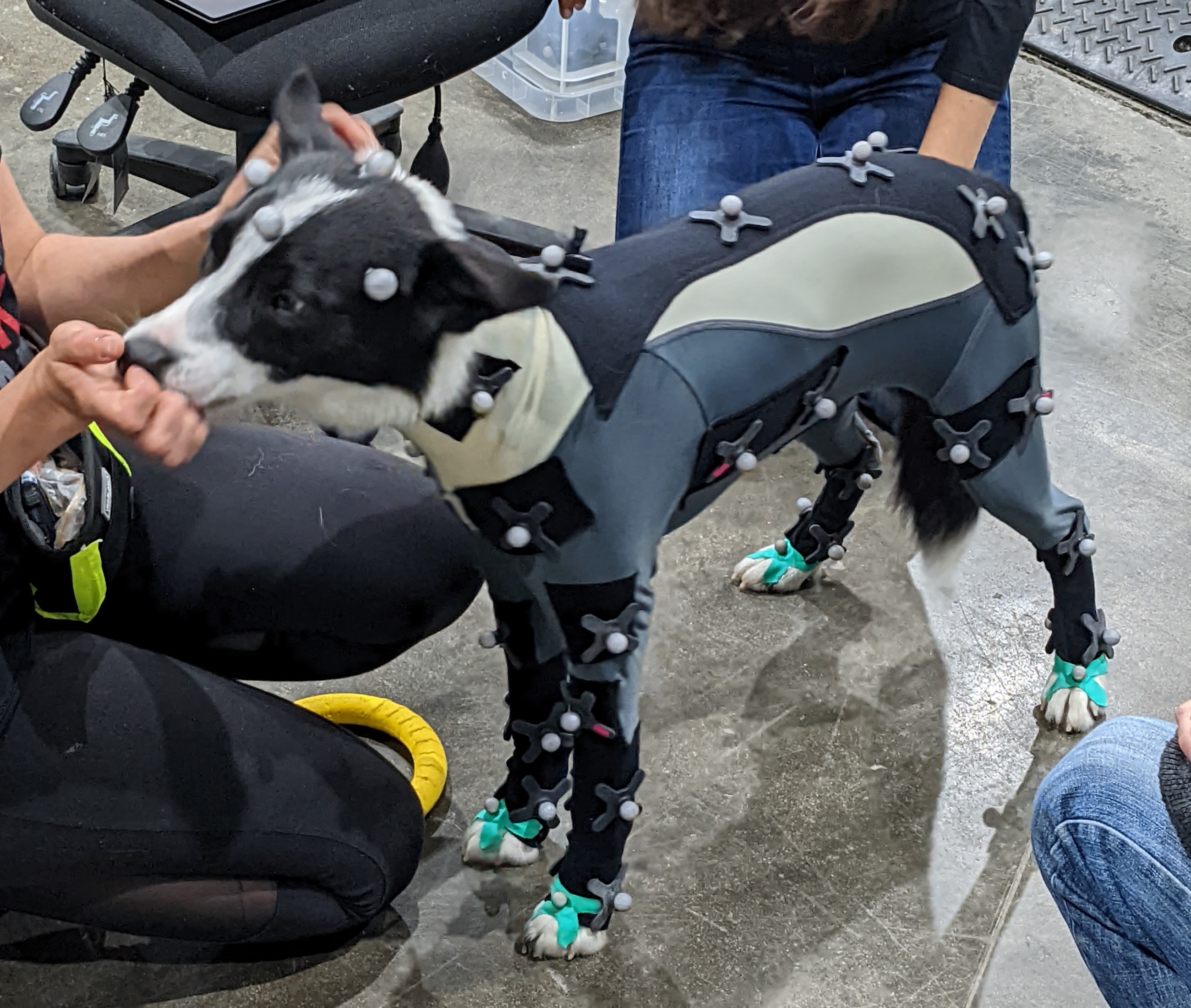}
    \includegraphics[width=0.3\linewidth]{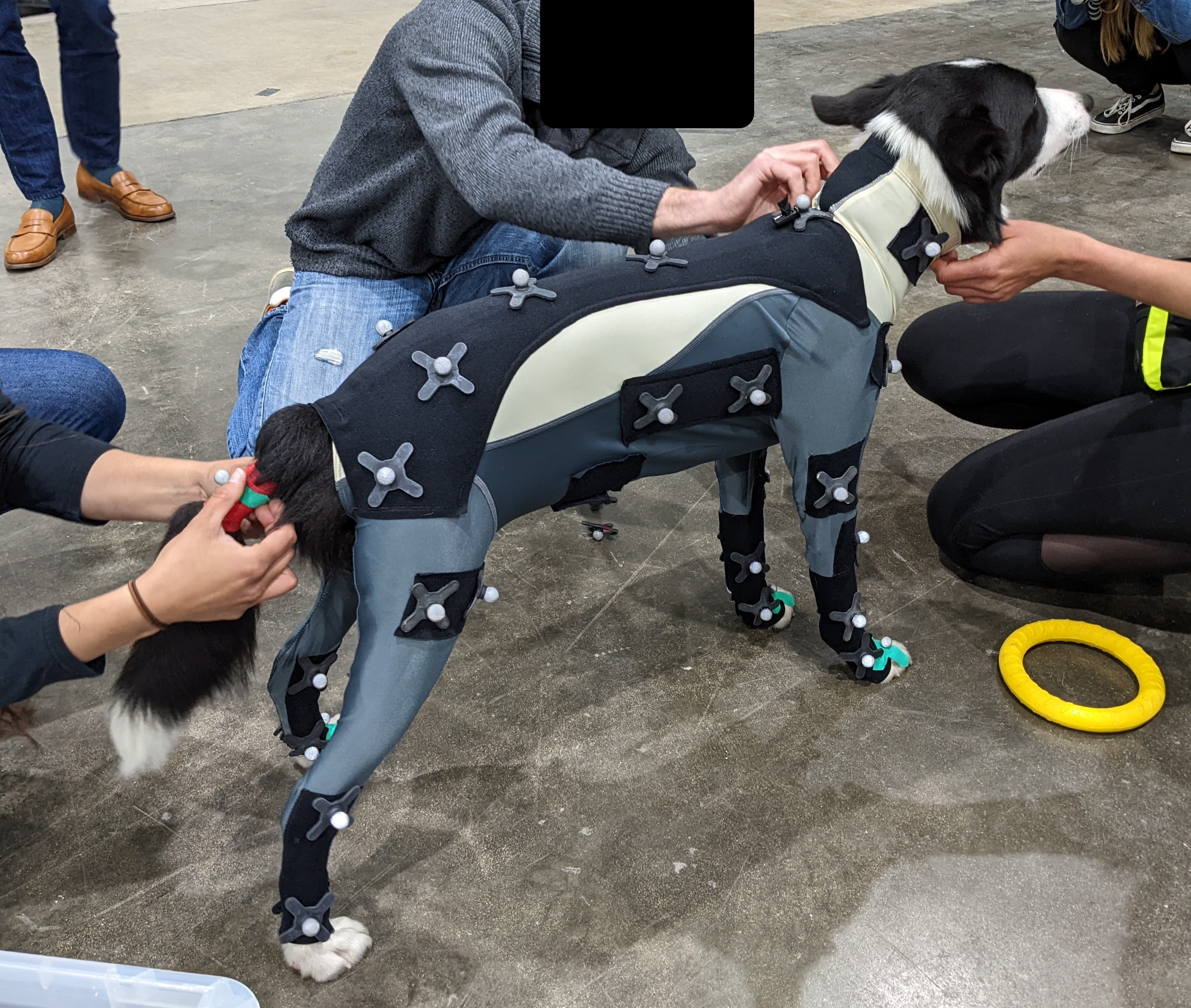}
    \includegraphics[width=0.3\linewidth]{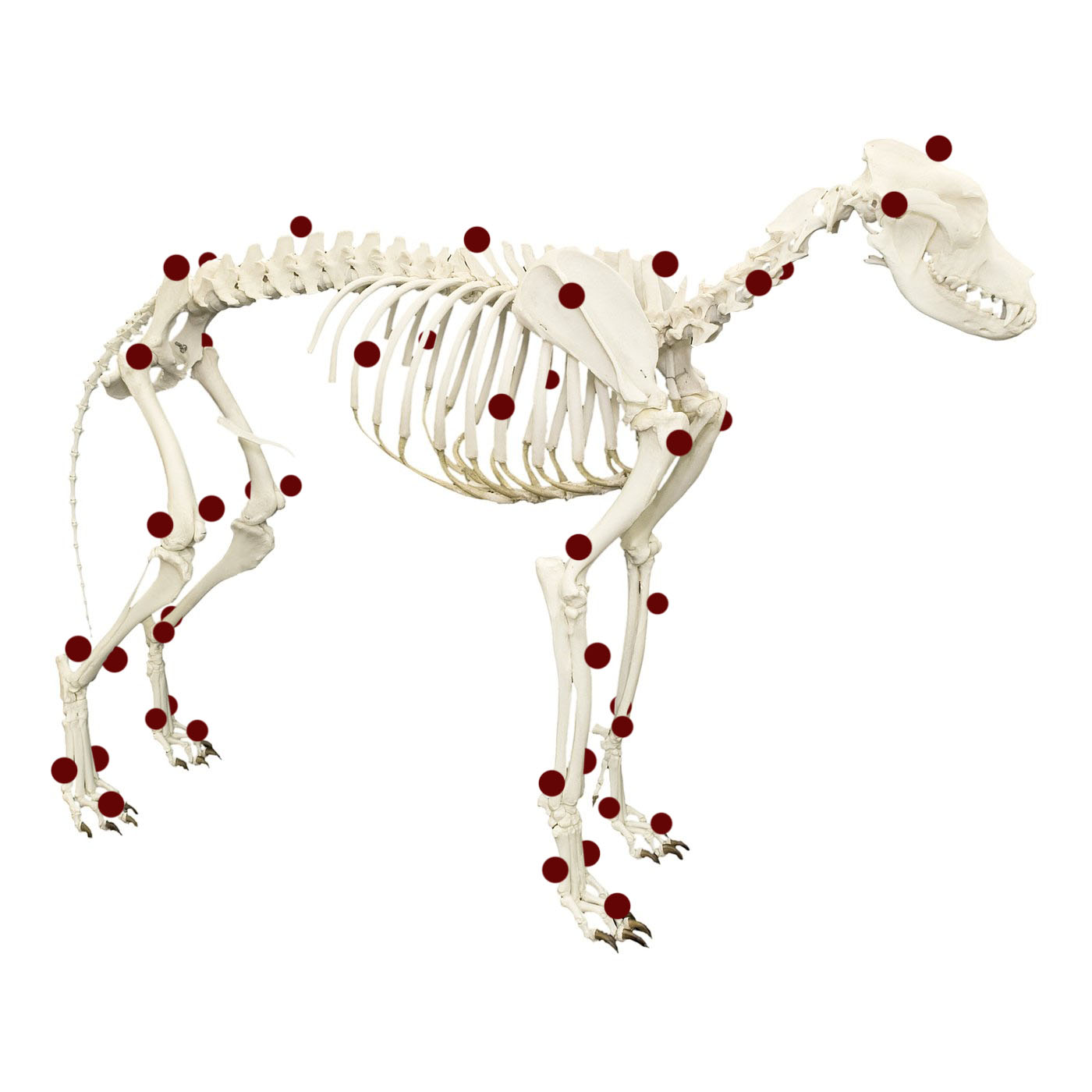}
    \caption{Marker positions used during motion capture. The markers are attached on a fitted suit which limits, but not entirely eliminates, their motion relative to their desired position, shown on the right.}
    \label{fig:markers}
\end{figure*}

Alongside this work we are releasing 28 clips of high-quality motion capture of a Border Collie. 
The selected clips capture a wide set of locomoting behaviours: walking, jumping, turning and sprinting.
The clips were obtained using a studio-grade, optical motion capture system, recording the accurate position of 53 reflective markers as shown in Fig.~\ref{fig:markers}.
The marker trajectories were post-processed after capture to assign unique labels and ensure continuity and completeness.

Although the dog model shares similarities with the one described in the previous section, morphological differences remain. Therefore, we need to retarget the motion capture data by converting 3D positions into joint angles for our skeleton.

We perform motion retargeting using an alternating optimization loop that iteratively refines both joint angles and motion capture scale. Within each iteration, we first optimize the joint angles: holding the current scale fixed, we follow \citet{buss2004introduction}, utilizing the Jacobian matrix and solving a damped least squares problem to minimize site position errors. This step ensures smooth movements while enforcing joint limits. Next, holding the newly adjusted joints fixed, we optimize the uniform scale of the motion capture data relative to the model using a bidirectional approach. This two-step process, alternating between joint optimization and scale optimization, is repeated until convergence is achieved.

After kinematic retargeting, we ensure that there aren't any penetrations with the ground by applying a physics-based post-processing step: we first disable gravity and then step each frame for 20ms simulated time with contact forces.
This ensures that each individual pose is sufficiently stable, without excessive forces.

\begin{figure*}[h!]
  \centering
  \includegraphics[scale=0.45]{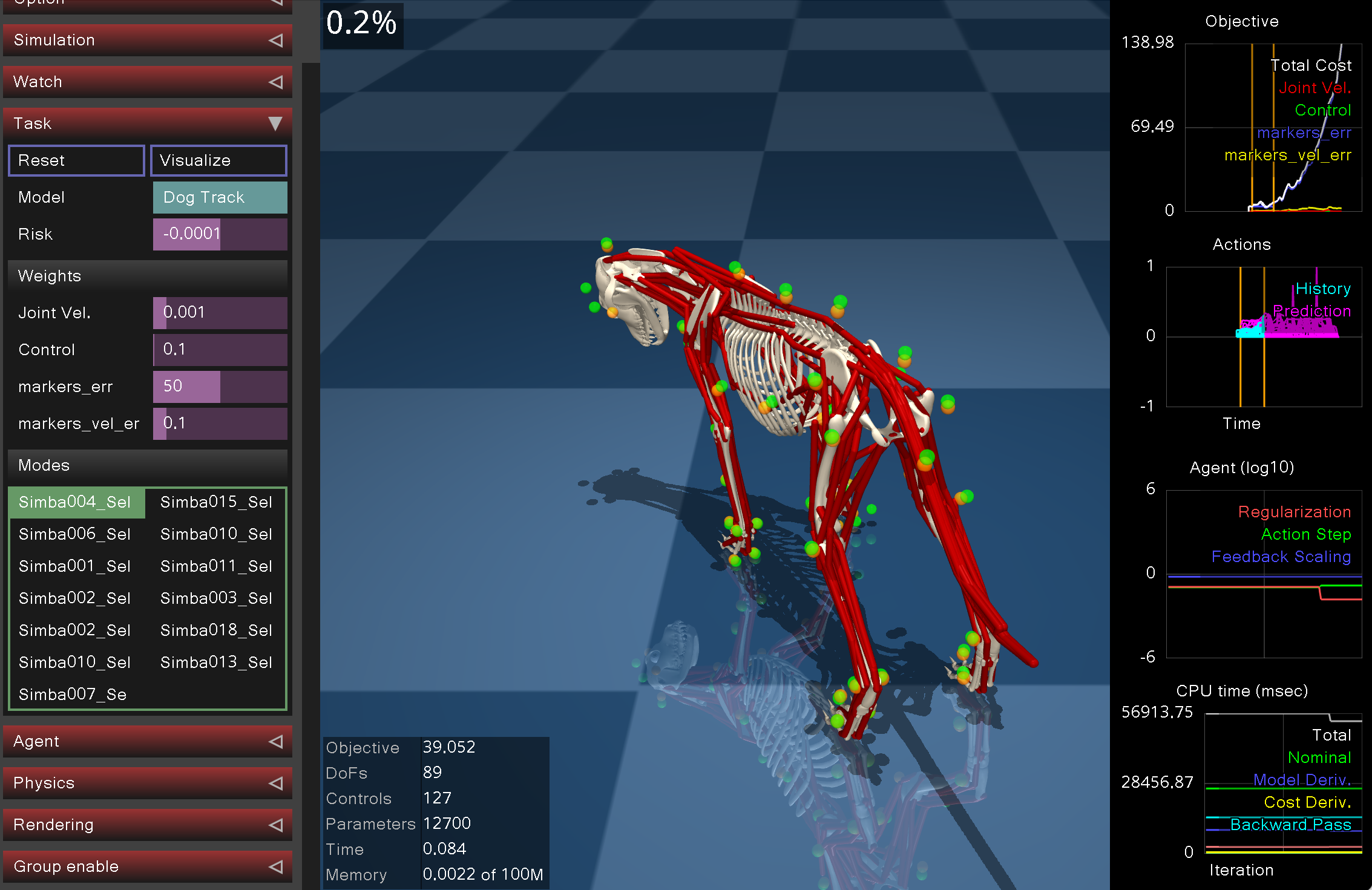}
  \caption{Graphical User Interface of MuJoCo MPC displaying the dog task.}
  \label{fig:dog_mjpc}
\end{figure*}

\subsection{Model Predictive Control}

\begin{figure*}[h!]
  \centering
  \includegraphics[scale=0.45]{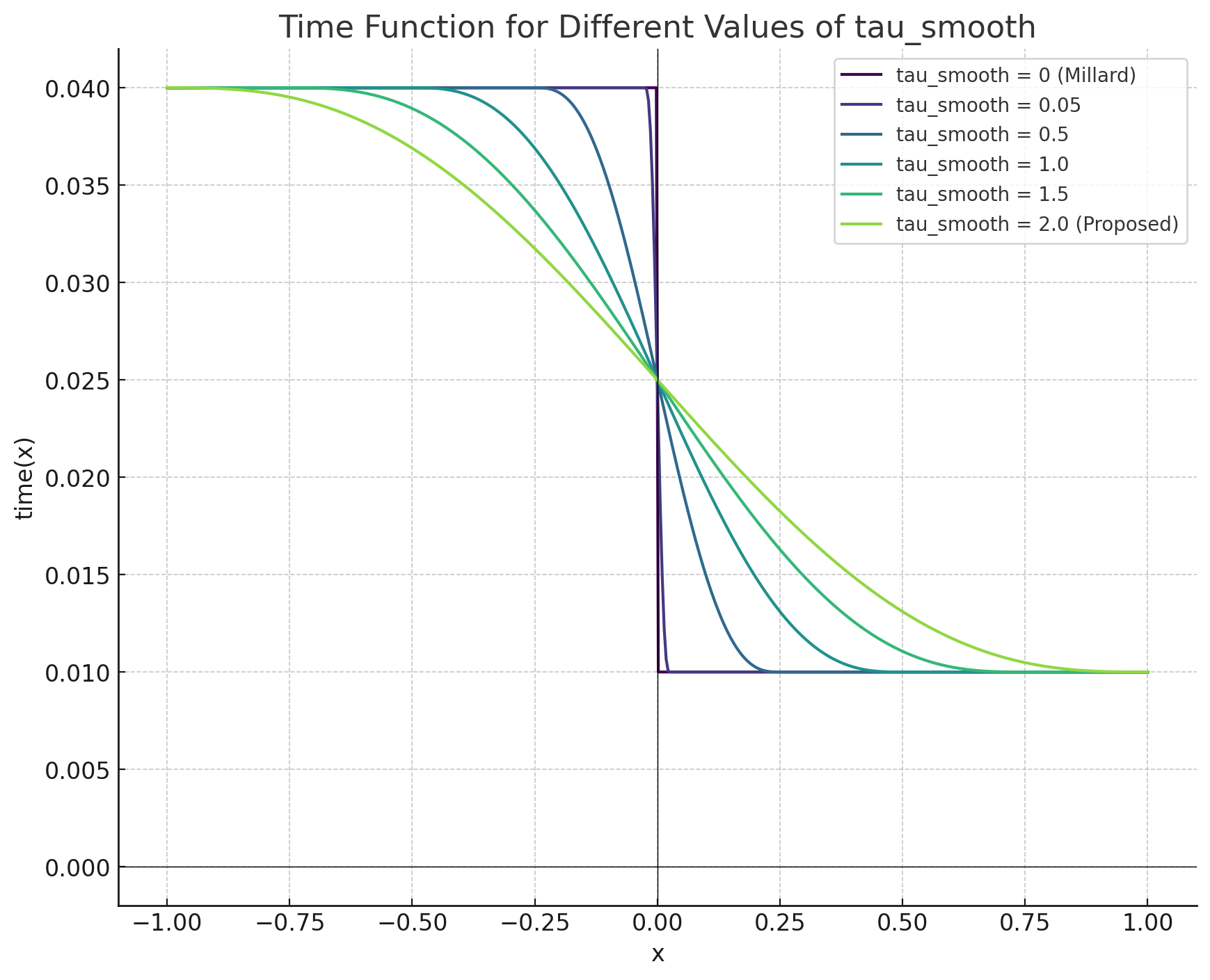}
  \caption[Sigmoid function used to replace the non-linear muscle dynamics proposed by Millard et al.]{The proposed sigmoid function for smooth transitions between activation and deactivation dynamics. As we can see as $\tau_{smooth}$ approaches  zero the closer we get to a hard switch between the time constants.}
  \label{fig:muscleDyn-sigmoid}
\end{figure*}

We tested the model capabilities with a motion capture tracking task. The goal of this task is to attempt to optimally track various locomotor behaviours and use these to estimate muscle functions (e.g., excitation patterns) for comparison with experimental data. One of the main contributions from this work is a modification in the muscle dynamics modelling, making them differentiable, which substantially increases the performance of the planning algorithm used to generate the control signals.

To control the model we chose a model predictive control algorithm implemented in MuJoCo Predictive Control (MJPC, \citet{howell2022}). MJPC was developed to ease the implementation of model-based optimisation, which often depends on highly-optimised and efficient algorithms and simulation models.
Specifically, we used the iLQG planner \citep{tassa2012Synthesis} for our tracking experiments. For details on the iLQG algorithm, we refer the reader to the references above.

In MPC the agent solves the problem by minimising an objective function that is designed as a weighted sum over costs:

\begin{equation}
   l(x, u) = \sum_{i=0}^{M} w_i \cdot n_i(r_i(x, u)) 
\end{equation}

Each of the $M$ terms comprises of:
\begin{itemize}
    \item A non-negative weight $w_i \in \Re$. Determining the relative importance of this term.
    \item A twice-differentiable norm $n_i(\cdot) : \Re^p \rightarrow \Re^{+}$, which takes its minimum at $0^p$ (zero-valued vector of size $p$).
    \item The residual $r_i \in \Re^p$ which is a vector of elements that are small when the task is solved.
\end{itemize}

The cost can then be customised with various norms and weights, and the details on the motion capture task are provided in the next section.

\paragraph{Tracking task} Similarly to the ostrich model presented by \citet{laBarbera21OstrichRL}, mocap data was used in this study to provide real-world reference for validating the musculoskeletal model and ensuring that the simulated movements align with actual locomotion. The use of mocap is necessary in both approaches to create accurate reward signals for reinforcement learning or to guide trajectory optimization, helping refine muscle actuation patterns and achieve realistic and biologically accurate gaits.

The motion tracking task is implemented following the cost previously defined. The terms used in the cost function are: 

\begin{itemize}
    \item joint velocities $r_0(x, u) = \dot{q}$, using a quadratic norm $n_0(r) = 0.5 \cdot r^2$. Aimed at penalising the joint velocities.

    \item muscle excitation (the control signal) $r_1(x, u) = u$, using a hyperbolic cosine norm $n_1(r) = p^2 \cdot (cosh(r / p) - 1)$ where $p=0.3$. Aimed at penalising the control signal; this is standard practice in simulations of musculoskeletal control to regularise muscle usage and avoid over-exerting individual muscles \citep{Anderson2001Dynamic, Thelen2006UsingCM, Dembia2021OpenSimMoco}.

    \item the positional error of the markers, defined as the difference between the Cartesian positions of the markers and their respective target positions $r_2(x, u) = m_{pos} - ref\_m_{pos}$. This term used a smooth absolute loss, $n_2(r) = \sqrt{r^2 + p^2} - p$ where $p = 0.1$, and aimed at providing a clear signal to the algorithm when it is drifting too much outside of the motion capture trajectory.
    
    \item the markers' velocities error, defined as the difference between the markers' linear velocities and their respective target velocities  $r_3(x, u) = m_{vel} - ref\_m_{vel}$. The same norm as for the positional markers' error was used but with a different scalar $p=0.3$.
    
\end{itemize}

We employ an alternative musculoskeletal model to overcome control challenges inherent in standard high-fidelity models. Specifically, the muscle dynamics proposed by \citet{millard2013flexing}, commonly used in MuJoCo, feature non-differentiable activation/deactivation delays that hinder gradient-based controllers like iLQG. Our chosen model uses differentiable dynamics while keeping the original tendon actuators, which we found to be effectively controllable.

Millard activation-deactivation dynamics are implemented in MuJoCo as follows:

\begin{equation}
    \tau(u, a) = \begin{cases} 
        \tau_{a} (0.5 + 1.5 a) & u > a \\
        \tau_{d} / (0.5 + 1.5 a) & u \leq a 
        \end{cases}
\end{equation}
We propose replacing the switching by a sigmoid (modulated by $\tau_{smooth}$), which smoothly interpolates between the two values within the range:

\begin{equation}
    time(x) =  \tau_{d} + (\tau_{a} - \tau_{d}) \cdot sigmoid(x/\tau_{smooth}  + 0.5)
\end{equation}

where $x = (u - a)$ is the difference between the control (excitation) and current activation. A plot showing the various curves of the sigmoid at the change of $\tau_{smooth}$ is shown in the figure  \ref{fig:muscleDyn-sigmoid}.
\begin{figure*}[h!]
  \begin{center}
      \includegraphics[scale=0.5]{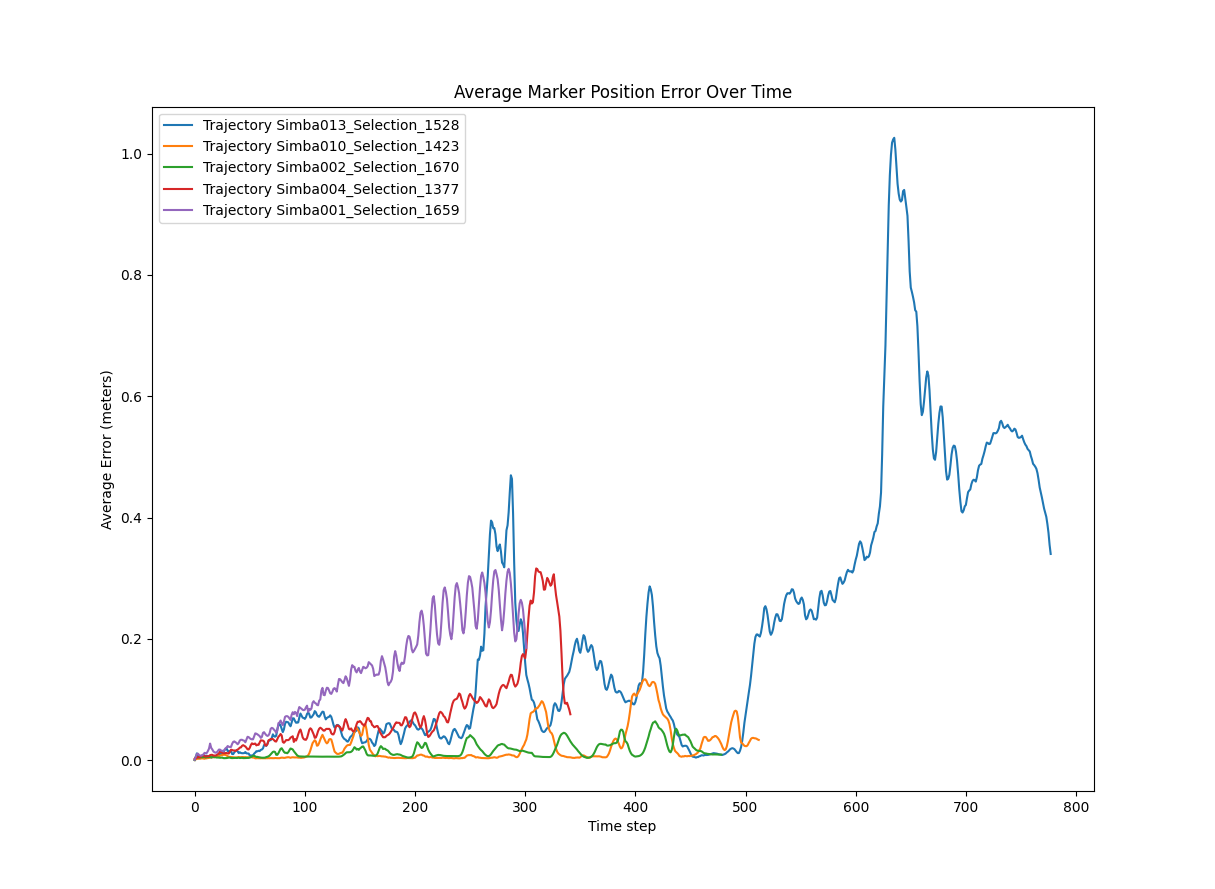}
      \caption[Dog simulations kinematic accuracy]{Graph showing kinematic accuracy of clips used to compare EMG}
      \label{fig:dog-kinematic-accuracy}
   \end{center}
\end{figure*}
\section{Results}

In this section we present results obtained by simulating the musculoskeletal dog model described in the previous chapter. The goal of the simulation is to attempt to optimally track various locomotor behaviours and use these to estimate muscle functions (e.g., excitation patterns) for comparison with experimental data. 
With the proposed muscle model change, iLQG is able to successfully track the selected motion capture clips.

Figure \ref{fig:dog-kinematic-accuracy} presents the kinematic error, which quantifies the deviation between the simulated and real marker positions across different motion clips. This error is computed as the average Euclidean norm of the difference between the simulated and ground truth marker positions:

\begin{equation}
    \text{Error}(t) = \frac{1}{M} \sum_{j=1}^{M} \left\| \mathbf{x}_{\text{sim}}^{(j)}(t) - \mathbf{x}_{\text{real}}^{(j)}(t) \right\|
\end{equation}

where $M$ is the total number of markers, $\mathbf{x}_{\text{sim}}$ represents the marker positions obtained from the simulation, and $\mathbf{x}_{\text{real}}$ represents the corresponding positions from the motion capture data. Across the evaluated motion clips, the error remains within 0.4 meters for walking, sprinting, and sit-to-stand movements, suggesting good tracking performance for these tasks. However, for the jumping motion, the error reaches up to 1 meter, indicating greater difficulty in accurately reproducing this dynamic movement. This discrepancy may stem from limitations in the muscle model's force production or inaccuracies in contact dynamics.

We will proceed and compare muscle excitation patterns to available EMG data in the literature. A lot of EMG data were available from the literature, and the following is a condensed list of what prior studies found.
We have selected 10 muscles (5 forelimb and 5 hindlimb) to compare against the simulations. 

The figures presented in this section show simulated EMG data generated from our musculoskeletal dog model. While a direct comparison with experimental canine EMG data would be ideal, the available literature is outdated and lacks publicly accessible datasets. Despite this limitation, our simulations produce activation patterns that align qualitatively with previous findings in quadrupedal locomotion research. Future studies should aim to validate these results with contemporary experimental EMG data.

\begin{figure*}[h!]
  \centering
  \includegraphics[scale=0.27]{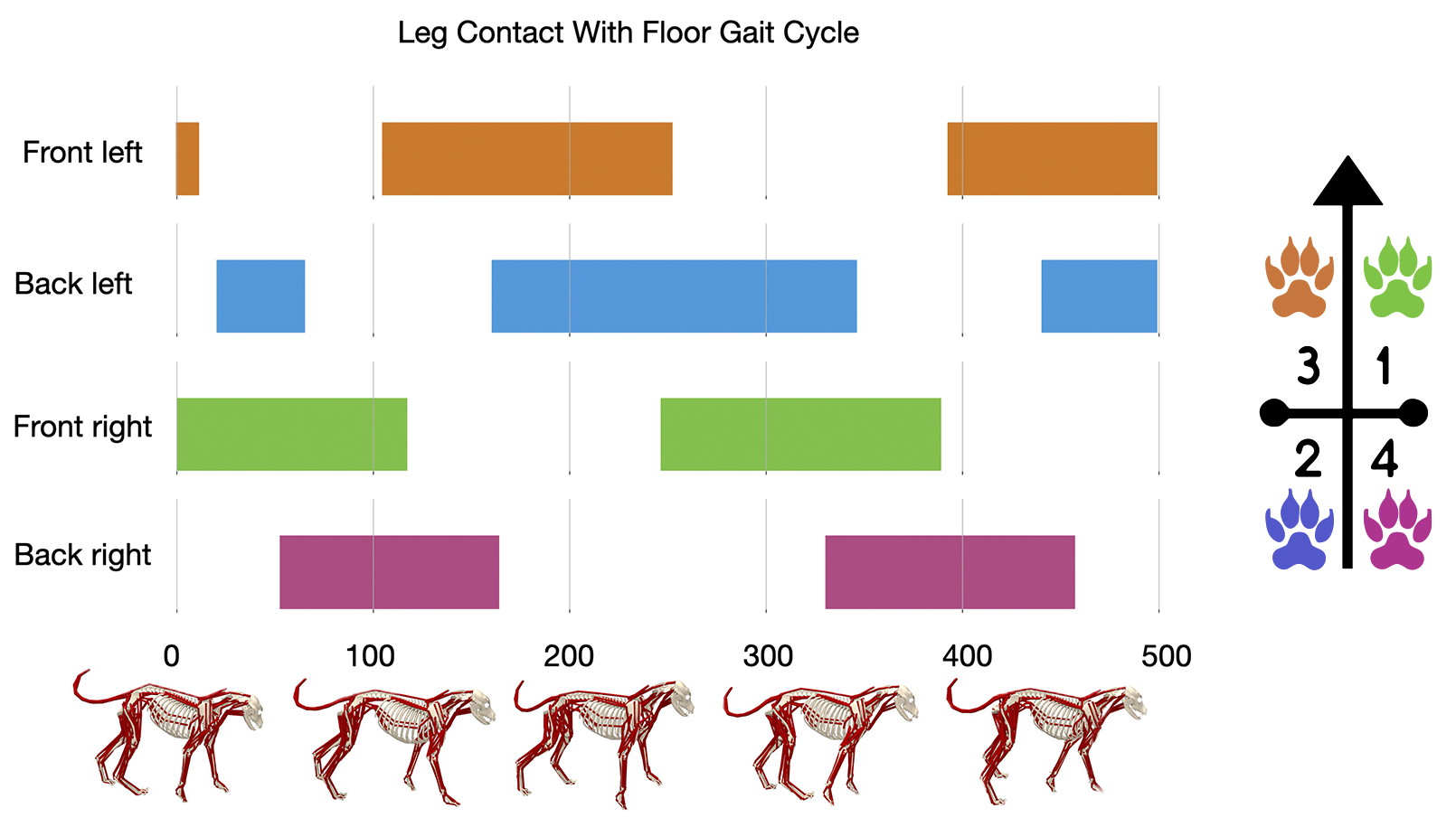}
  \caption[Foot contacts of the musculoskeletal dog during walking]{The foot contact patterns observed in simulation when the dog is walking}
  \label{fig:dog-foot-contact}
\end{figure*}

We compare 3 actions from EMG data: walking, sit to stand and jumping.
To compare the EMG data for walking, we used a full walking gait cycle from one of the clips. Note that the recorded motions were not acquired in a biomechanics lab but rather in a motion capture studio and that the purpose was to record a wide range of motions rather than studying controlled movements of the dog like walking straight on a treadmill.
The foot contact placement for a walking straight clip can be visualised in the Fig.~\ref{fig:dog-foot-contact}. This diagram will be used to observe swing and stance phases of the various legs in the EMG comparison.

\paragraph{Walking: Hindlimb Muscles}

\emph{Biceps Femoris \& Semimembranosus (Deban et al., 2012, Goslow et al.):} These muscles, primarily responsible for hindlimb retraction, were active during the latter part of swing and the initial part of stance, suggesting a role in both decelerating the swinging limb and contributing to propulsion during stance. The increased activity observed in galloping aligns with the need for greater propulsive forces at higher speeds. The biceps femoris also plays a role in knee flexion, particularly around touchdown and liftoff.

\begin{figure*}[h!]
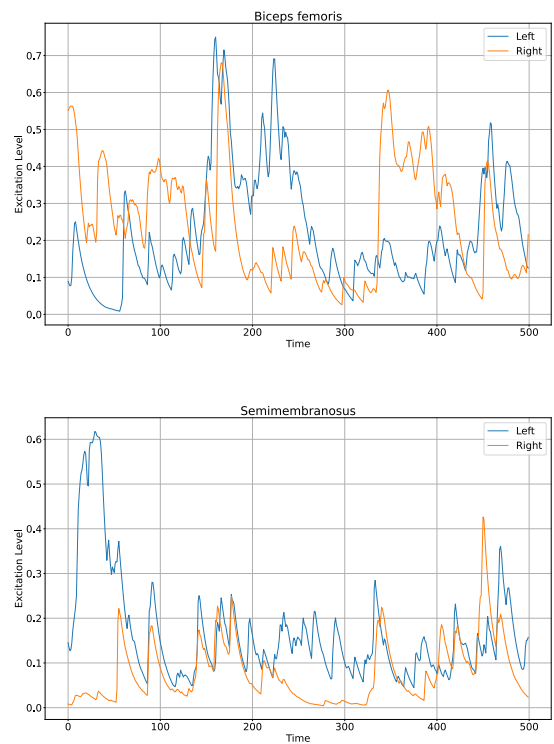

      \centering
      \includesvg[scale=0.17]{figures/EMG_plots/biceps_femoris.svg}
      \includesvg[scale=0.17]{figures/EMG_plots/semimembranosus.svg}
      \caption{Muscle Excitation of Biceps femoris and Semimembranosus}
      \label{fig:dog-activation-biceps-femoris}
\end{figure*}

The two plots in Fig.~\ref{fig:dog-activation-biceps-femoris} from simulation data show accordance with the biceps femoris being active mostly during the swing phase for the right and left hindlimb. However, the semimembranosus shows little activity through out the movement, making it impossible to distinguish swing or stance variations.

\begin{figure*}[h!]
      \centering
      \includesvg[scale=0.17]{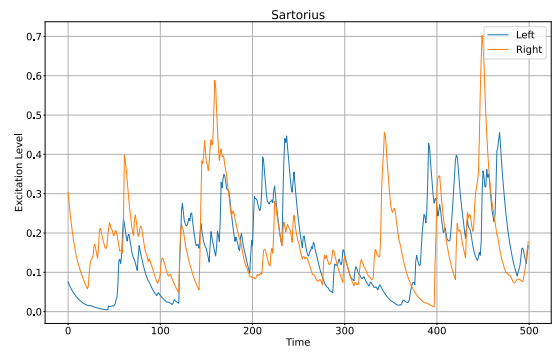}
      \caption{Muscle Excitation of Sartorius}
      \label{fig:dog-activation-sartorius}
\end{figure*}

\emph{Sartorius (cranial \& caudal) (Deban et al., 2012; Goslow et al., 1981):} From the literature, the sartorius muscles are primarily active during the swing phase, contributing to hip flexion and limb protraction. The caudal sartorius showed peak activity in trotting, potentially due to the rapid swing phase in this gait. The cranial sartorius was more active during walking and galloping, suggesting a more complex role in limb protraction that varies with gait.
In the plot obtained from simulation Fig.~\ref{fig:dog-activation-sartorius}, there are similar excitation patterns in the right leg where the muscles are active during the swing phases; while for the left leg the only clear signal is during the second swing phase (at the $400^{th}$ frame).

\begin{figure*}[h!]
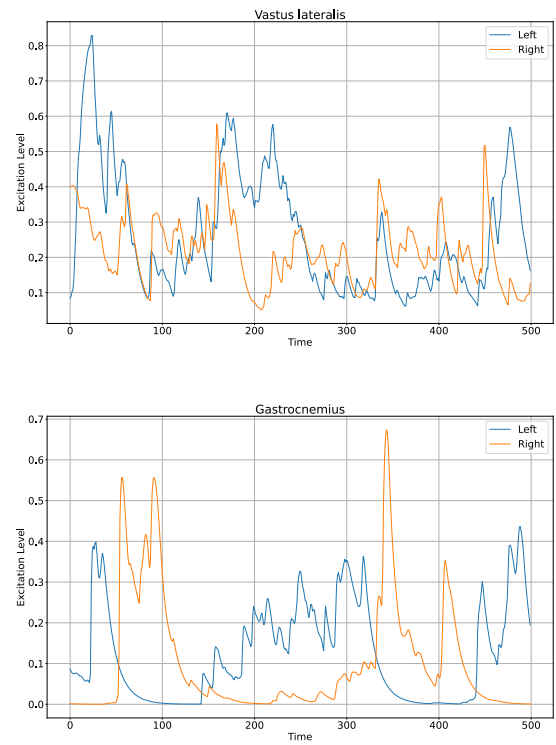

    \centering
    \includesvg[scale=0.17]{figures/EMG_plots/vastus_lateralis.svg}
    \includesvg[scale=0.17]{figures/EMG_plots/gastrocnemius.svg}
    \caption{Muscle Excitation of Vastus Lateralis and Gastrocnemius}
    \label{fig:dog-activation-vastus}
\end{figure*}

\emph{Vastus Lateralis \& Gastrocnemius (Goslow et al., 1981):} As hindlimb extensors, these muscles exhibited activity patterns consistent with a spring-like mechanism, particularly in trotting and galloping. They undergo both stretching and shortening during stance, suggesting a role in energy storage and recovery.
In the plots obtained from simulation in Fig.~\ref{fig:dog-activation-vastus} we can observe very clear excitation patterns in the gastrocnemius, being active during stance phase and complementing nicely between the right and left legs.
The vastus lateralis on the other hand shows almost constant excitations during the various gait phases, preventing meaningful conclusions.

\begin{figure*}
     \centering
    \includegraphics[width=0.4\linewidth]{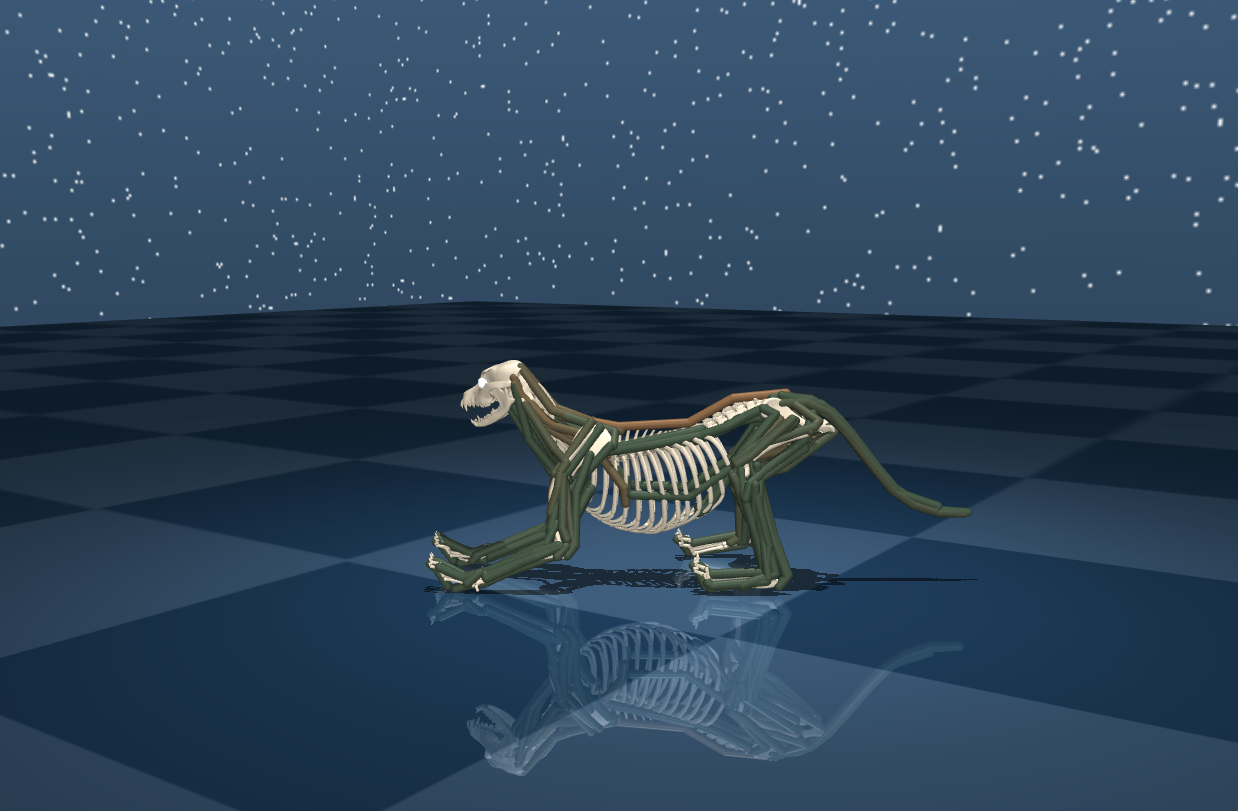}
    \includegraphics[width=0.4\linewidth]{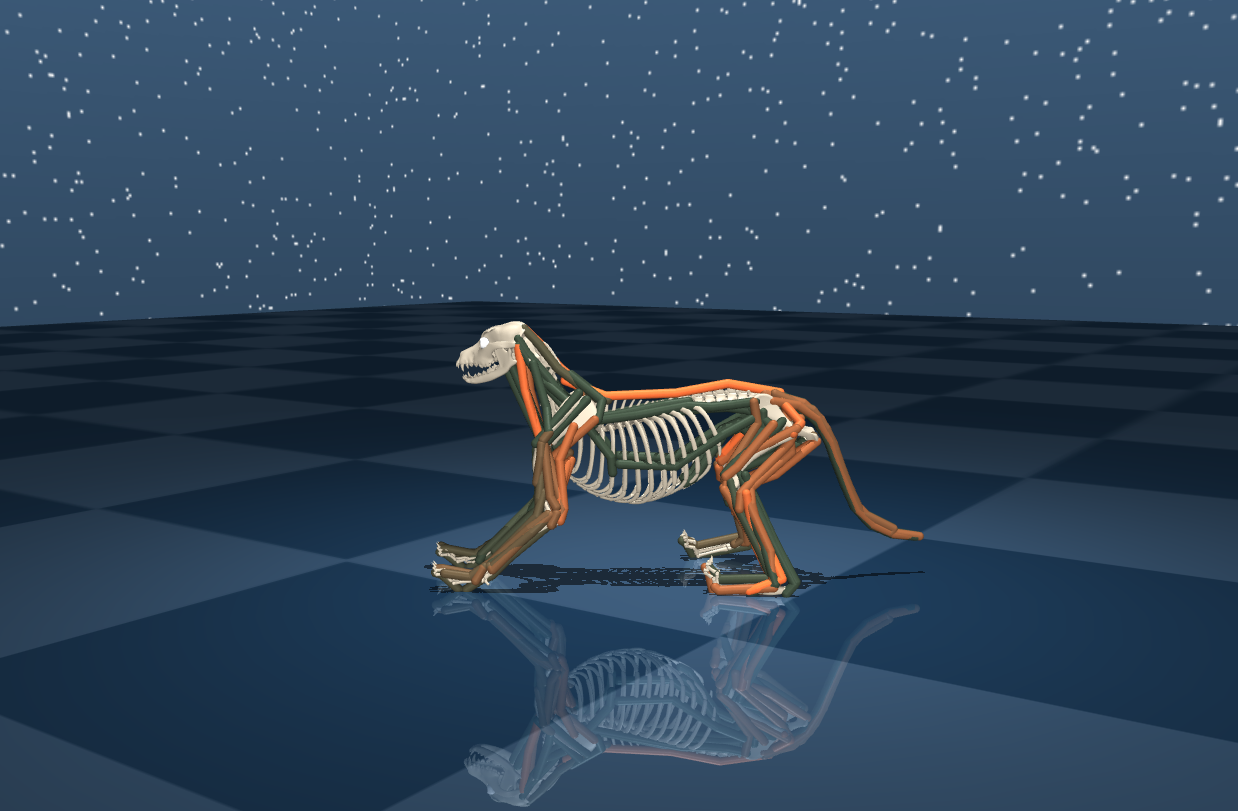}
    \includegraphics[width=0.4\linewidth]{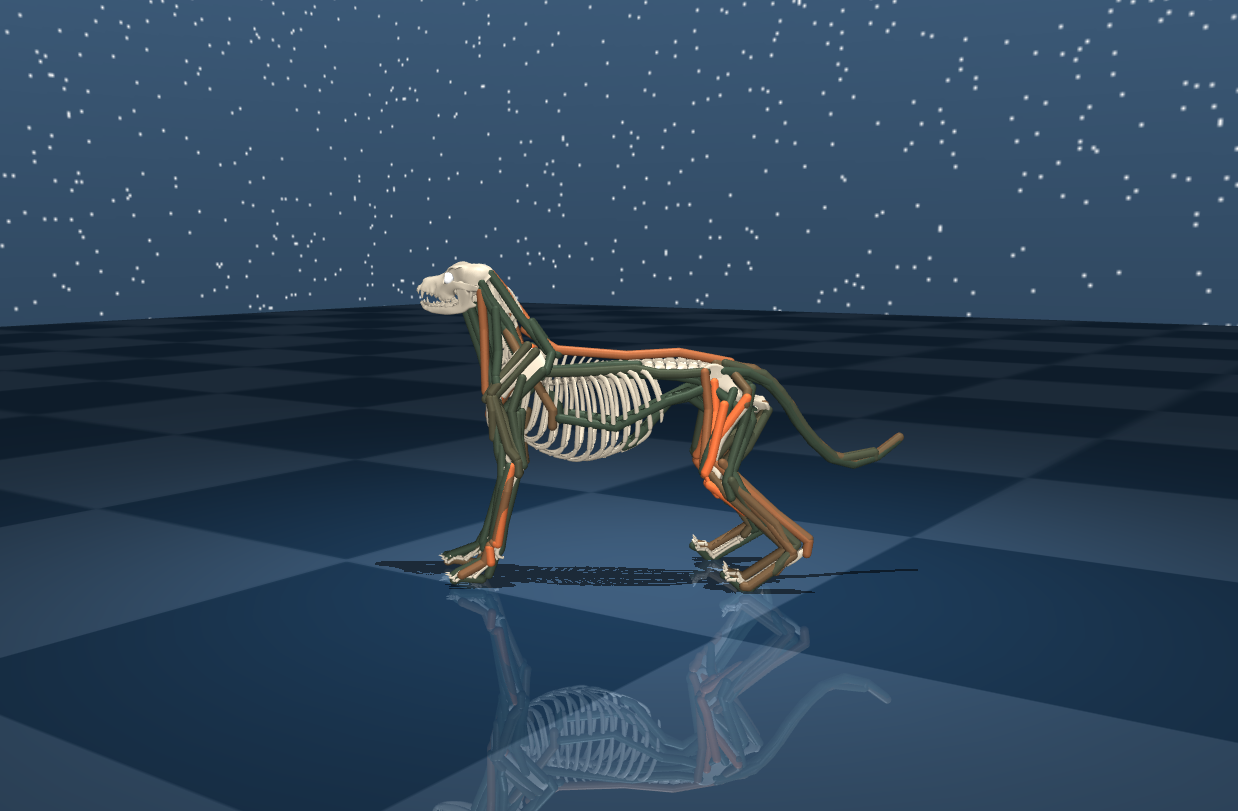}
    \includegraphics[width=0.4\linewidth]{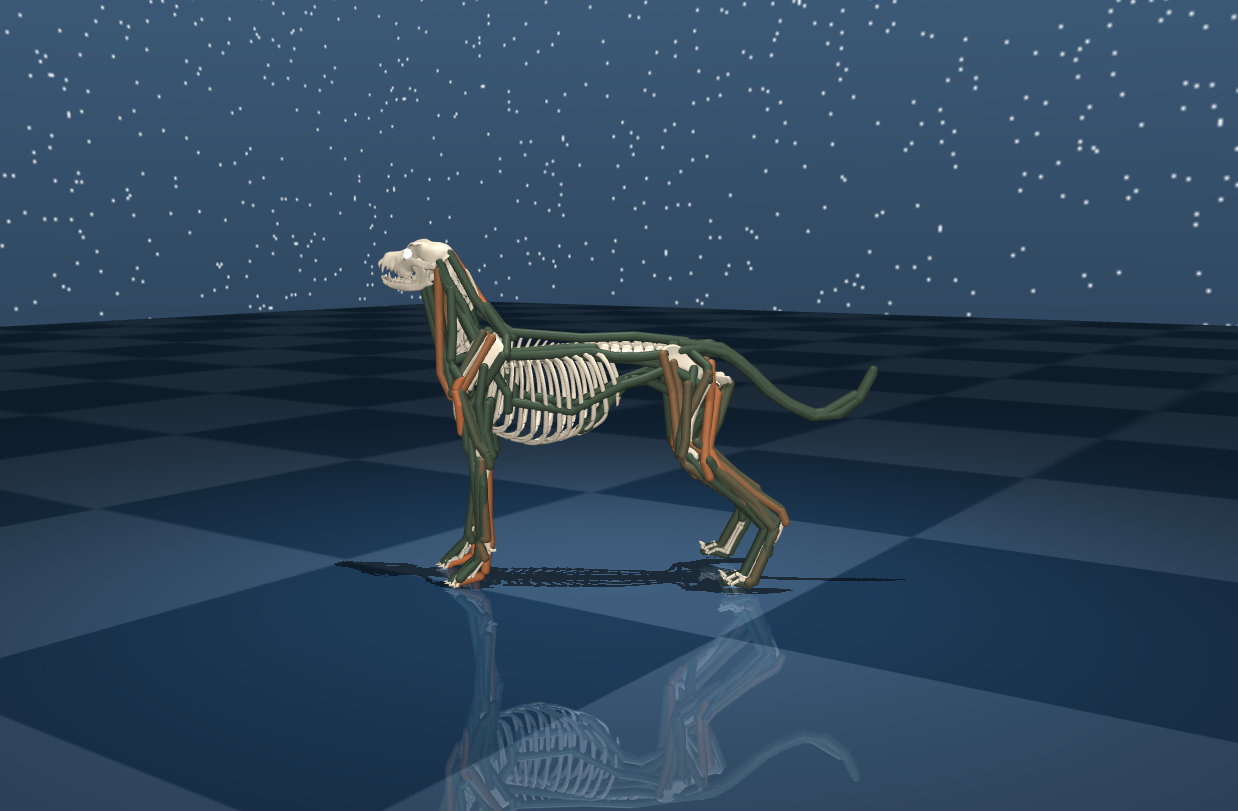}
    \caption[Sit to Stand transition in simulation for the musculoskeletal dog model]{Sit to Stand transition in simulation for the musculoskeletal dog model. Transition starts at the top left and ends standing at the bottom right.}
    \label{fig:dog-sts}
\end{figure*}

\paragraph{Walking: Forelimb Muscles}

\begin{figure*}[h!]
      \centering
      \includesvg[scale=0.17]{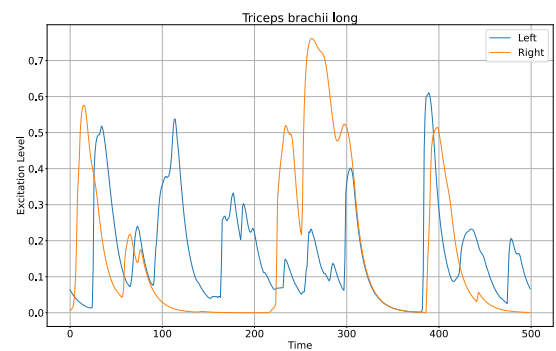}
      \caption{Muscle Excitation of Triceps Brachii}
      \label{fig:dog-activation-tricepsbrachii}
\end{figure*}

\emph{Triceps Brachii (long \& lateral heads) (Cullen et al., 2017; Goslow et al., 1981):} From the literature it is evident that the triceps brachii is a major extensor of the elbow joint. In walking, the lateral head is inactive, while the long head is active during late swing and early stance, likely contributing to deceleration of the forelimb before foot contact and initial weight-bearing. In trotting and galloping, both heads show increased activity duration, reflecting a greater role in elbow extension and energy management at faster speeds. The long head, being biarticular, also contributes to shoulder flexion, particularly during the demanding jumping task.
Observing the plot obtained from simulation in Fig.~\ref{fig:dog-activation-tricepsbrachii}, the muscle is mostly active during the stance phase, which is in accordance with EMG data suggesting a contribution to deceleration of the forelimb before foot contact.

\begin{figure*}[h!]
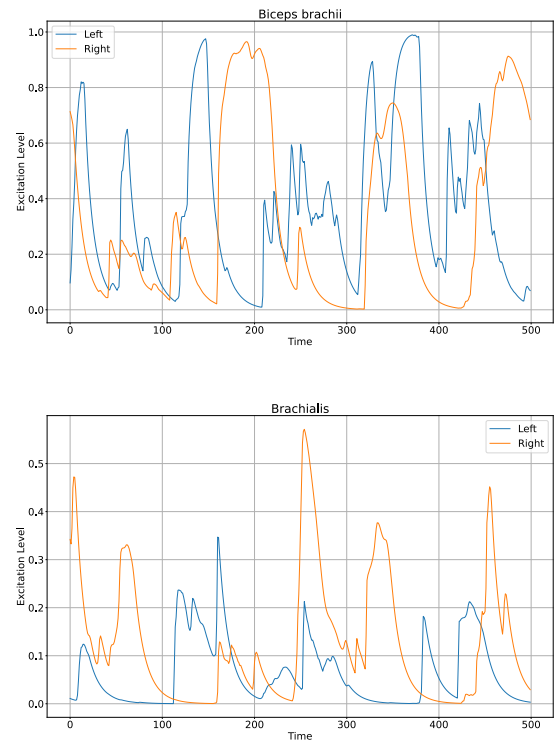

    \centering
    \includesvg[scale=0.17]{figures/EMG_plots/biceps_brachii.svg}
    \includesvg[scale=0.17]{figures/EMG_plots/brachialis.svg}
    \caption{Muscle Excitation of Biceps Brachii and Brachialis}
    \label{fig:dog-activation-biceps}
\end{figure*}

\emph{Biceps Brachii \& Brachialis (Cullen et al., 2017; Goslow et al., 1981):} These muscles are primarily elbow flexors. In walking, they are active mainly during the swing phase, contributing to limb protraction. In trotting and galloping, their activity is concentrated in the early swing phase, suggesting a role in rapid elbow flexion and limb acceleration at the beginning of swing. The biceps brachii, being biarticular, also acts as a shoulder flexor and shows high excitation during jumping, similar to the triceps brachii long head.
In the plots Fig.~\ref{fig:dog-activation-biceps} obtained from the simulation, the brachialis is more active during the swing phases while the biceps brachii is active in the late stance phases and initial swing phases.

\begin{figure*}
     \centering
  \includegraphics[scale=0.25]{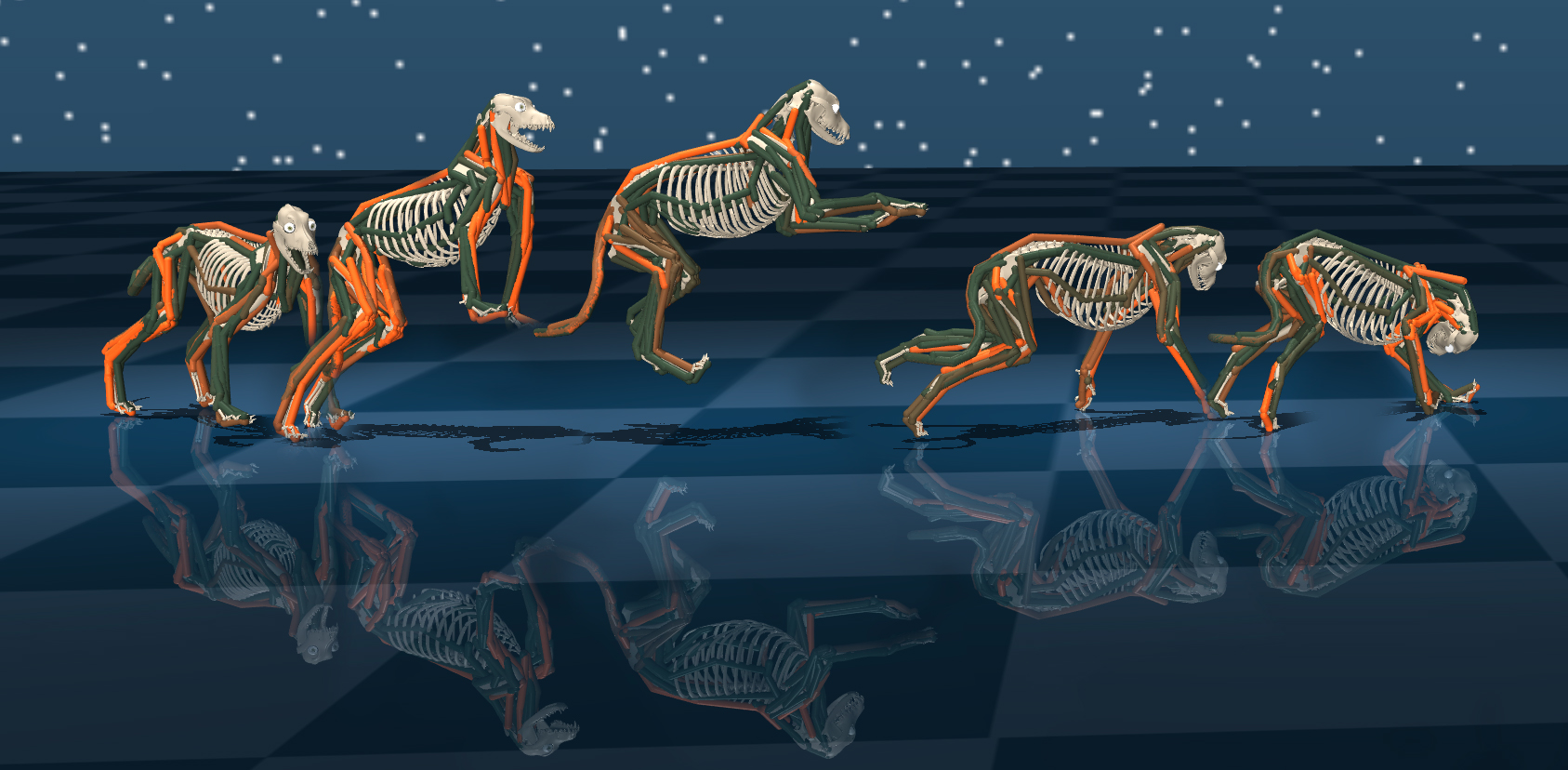}
  \caption[A capture of a dog jumping during simulation]{The proposed musculoskeletal model successfully tracking a jumping clip. The muscle excitations are shown using a gradient from green (inactive) to orange (fully excited).}
  \label{fig:dog-jumping}
\end{figure*}

\paragraph{Sit to stand transition}

We also compare sit-to-stand (StS) transitions with what has been previously
found in the literature. \cite{Ellis2018Limb} showed that several muscles, including the gluteus superficialis, adductor magnus, knee extensors (vastus lateralis, vastus medialis, rectus femoris), and ankle extensors (flexor digitorum superficialis, gastrocnemius lateralis and medialis) consistently reached maximal activation (100\%) during StS. This highlights their critical role in generating the necessary joint torques to overcome the challenges of transitioning from a crouched to an upright posture.
The muscle excitations found in simulation are shown below in figure \ref{fig:dog-activations-sts}. Important time windows in the plots are: From the 0 to 100 frame the dog is sitting still. From the 100 to 300 frame the dog is transitioning to standing up. From 300 to 700 the dog is standing still and from 700 to 1000 is sitting back down.
From the plots we see clearly that the adductor and the vastus lateralis are active during the standing up and sitting down transitions, which is in agreement from the experiments conducted by \cite{Ellis2018Limb}. However the flexor digitorum superficialis shows little constant excitation through out the entire motion, without any clear peaks during the motion. The sit to stand transition in simulation is shown in Fig.~\ref{fig:dog-sts}.

\paragraph{Jumping}
Lastly we compare muscle excitations during jumping. Fig.~\ref{fig:dog-jumping} shows the musculoskeletal model successfully tracking a jumping clip. Below there are the plots of four muscles excitations which can be compared with data from the literature. The dog was airborne from frame 70 to frame 210. 
The triceps brachii and the infraspinatus were most active during take off and landing which is in agreement with the findings in the literature. While the supraspinatus was most active while the dog was mid-air, this is also in agreement with previous data showing that the supraspinatus is less involved in propulsive phases.
The biceps brachii was very active throughout the task. All of these plots very clearly show that all of the selected muscles show peak excitations that are much higher than during walking, highlighting the physical demand this task poses on the dog.

\section{Conclusions}
We demonstrated that the proposed model, controlled via the iLQG implementation in MJPC, effectively tracks complex motion capture sequences, highlighting the robustness of this algorithm. Our findings further indicate that modifying the model to include differentiable muscle dynamics is essential for accurately tracking these motion capture clips.

When comparing simulated muscle excitations with previously reported EMG data, we observed substantial agreement across a range of movements. However, specific exceptions were noted: the vastus lateralis and infraspinatus muscles during walking, the flexor digitorum superficialis during sit-to-stand transitions, and notably high activity in the biceps brachii during jumping.

Additionally, as previously mentioned, these motion capture sequences were recorded in less controlled conditions compared to treadmill-based settings, which naturally leads to imperfect matches in simulation results. Lastly, morphological differences between the dog breed used in the motion capture recordings and the simulated model should also be acknowledged as contributing to observed discrepancies.

\bibliographystyle{abbrvnat}
\nobibliography*
\bibliography{template_refs}

\section*{Funding}
This research was funded by DeepMind.

\section*{Competing interests}
The authors declare no competing financial interests. Related
patent number here if applicable.

\section*{Data availability}
The model will be made available at the following GitHub repository: \url{https://github.com/vittorione94/MusculoskeletalDog}.

\section*{Supplementary Materials}

\subsection{Line of action estimation}
\begin{algorithm}
\caption{Line of action estimation}\label{euclid}
    \begin{algorithmic}[1]
    \Procedure{LOA}{$skeleton, mesh, axis, samples$}
    
    \State $tendon\gets []$
    \State $KDTrees\gets \{\}$
    \For{$bone\in skeleton$}
    \State $KDTrees[bone.name] \gets kdTree(bone)$
    \EndFor
    
    \State $contours\gets slice(mesh, axis, samples)$
    \State $centroids\gets [] $
    \For{$c\in contours$}
    \State $centroids.append(c.centroid())$
    \EndFor
    
    \State $addedSites \gets []$
    
    \For{$c\in centroids$}
    \State $geom \gets getClosestGeom(KDTrees, c)$
    \State $d \gets distance(c, addedSites.last)$
    
    \State $conds \gets False$
    \State $cond$ {\bf or} $addedSites.empty()$
    \State $cond$ {\bf or} $c.last$
    \State $cond$ {\bf or} $d \geq maxDist$
    \State $cond$ {\bf or} ($newBody$ {\bf and} $d \geq minDist$)
    \If{$cond$}
        \State $tendon.addSite(c)$
        \If{$wrapping$}
            \State $tendon.addWrapping()$
        \EndIf
    \EndIf
    
    \EndFor
    
    \State \textbf{return} $tendon$
    \EndProcedure
    \end{algorithmic}
    
    \label{line of action}
\end{algorithm}

\subsection{Muscle excitation plots from simulation}

All excitation plots in this work are defined with excitation signals that are dimensionless, as they represent a normalized excitation level that ranges from 0 to 1. A value of 0 indicates that the muscle is fully relaxed, while a value of 1 indicates that the muscle is maximally excited, meaning it is receiving full neural input.

\begin{figure*}[h!]
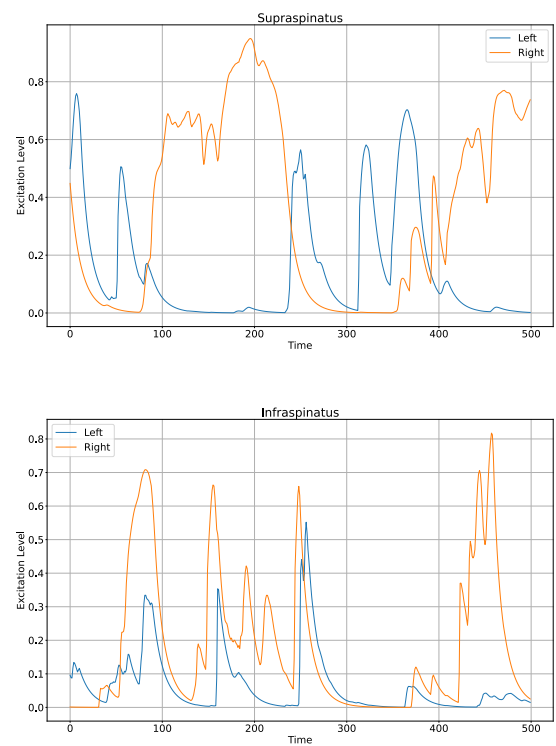

    \centering
    \includesvg[scale=0.17]{figures/EMG_plots/supraspinatus.svg}
    \includesvg[scale=0.17]{figures/EMG_plots/infraspinatus.svg}
    \caption{Muscle Excitation of Supraspinatus and Infraspinatus}
    \label{fig:dog-activation-infraspinatus}
\end{figure*}

\begin{figure*}[h!]
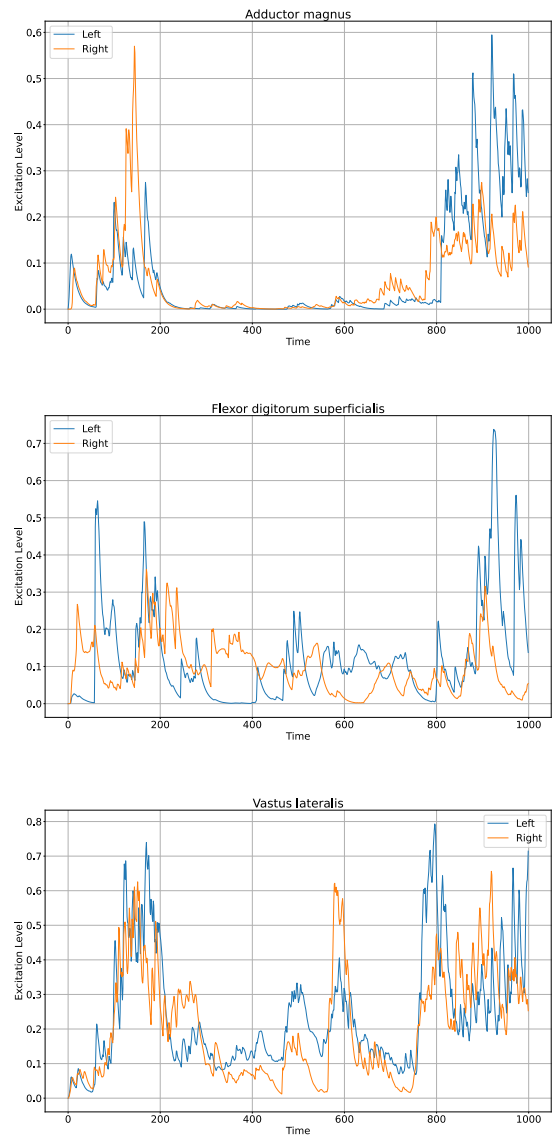

      \centering
      \includesvg[scale=0.17]{figures/EMG_plots/adductor_sts.svg}
      \includesvg[scale=0.17]{figures/EMG_plots/flexor_digitorum_superficialis_sts.svg}
      \includesvg[scale=0.17]{figures/EMG_plots/vastus_lateralis_sts.svg}
      \caption[Excitation patterns of three muscles acquired during simulation to analyse the sit to stand transition in dogs]{Excitation patterns of three muscles (adductor, flexor digitorum superficialis, vastus lateralis) acquired during simulation to analyse the sit to stand transition in dogs.}
      \label{fig:dog-activations-sts}
\end{figure*}

\begin{figure*}
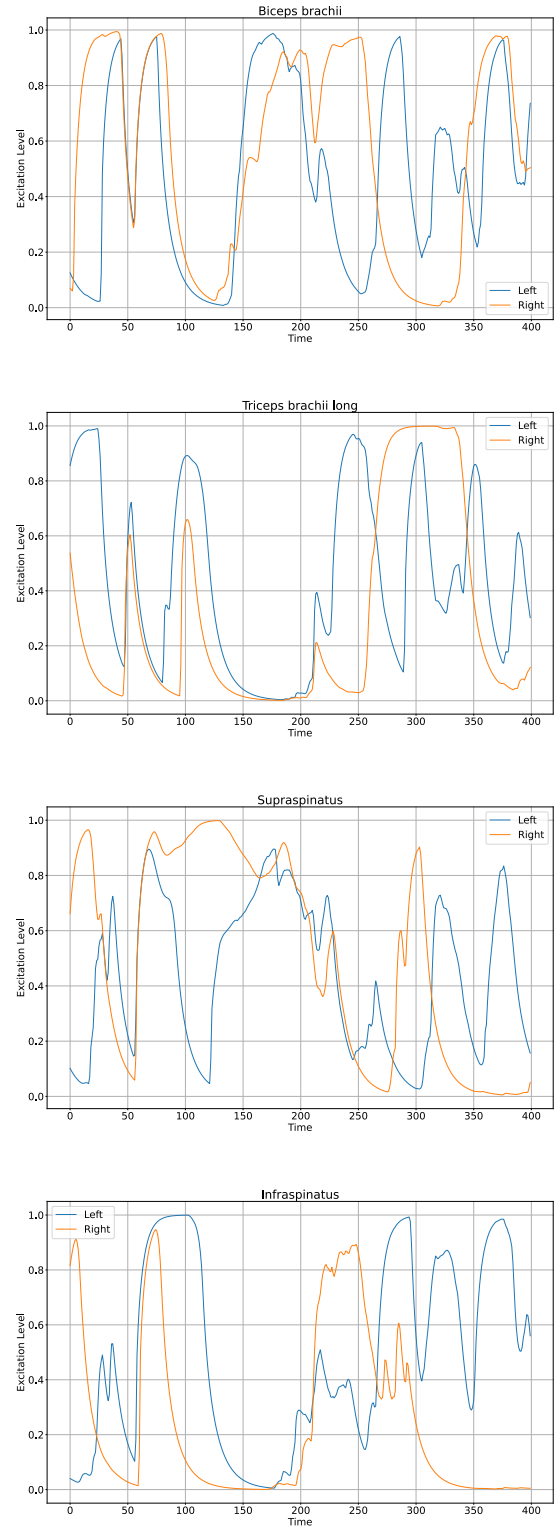

    \centering
    \includesvg[scale=0.17]{figures/EMG_plots/biceps_brachii_jump.svg}
    \includesvg[scale=0.17]{figures/EMG_plots/triceps_brachii_jump.svg}
    \includesvg[scale=0.17]{figures/EMG_plots/supraspinatus_jump.svg}
    \includesvg[scale=0.17]{figures/EMG_plots/infraspinatus_jump.svg}
    \caption[Excitation patterns of four muscles acquired during jumping in  simulation]{Excitation patterns of four muscles (biceps brachii, triceps brachii, supraspinatus, infraspinatus) acquired during jumping in simulation.}
    \label{fig:dog-activations-jumping}
\end{figure*}

\end{document}